%% file: main.tex
\documentclass[letterpaper,journal]{IEEEtran}

\usepackage[linesnumbered,ruled,vlined]{algorithm2e}

\usepackage{amssymb}
\usepackage{algorithmicx}

\usepackage{lineno}
\usepackage{subfig}
\usepackage{booktabs,caption}
\usepackage[flushleft]{threeparttable}
\usepackage{amsfonts}
\usepackage{float}
\usepackage{placeins}
\usepackage{textcomp}
\usepackage{amsmath}
\usepackage{graphicx}
\usepackage{url}
\usepackage{multicol, blindtext}
\usepackage{booktabs}
\usepackage[table]{xcolor}
\usepackage{xcolor}
\usepackage{algorithmicx}
\usepackage{lipsum}
\usepackage{amssymb}
\usepackage{lineno}
\usepackage{algpseudocode}
\usepackage{amsfonts}
\usepackage{siunitx}
\usepackage{amsmath}
\usepackage{multirow}
\usepackage{wrapfig}
\usepackage{array}
\usepackage{color}
\usepackage{silence}
\usepackage[table]{xcolor}
\definecolor{lightgray}{gray}{0.9}
\usepackage{subcaption}

\usepackage{tabularx}

\usepackage{amssymb}
\usepackage{lineno}
\usepackage{subfig}
\usepackage[english]{babel}
\usepackage{ragged2e}
\usepackage{blindtext}
\usepackage{booktabs,caption}
\usepackage[flushleft]{threeparttable}
\usepackage{amsfonts}
\usepackage{float}
\usepackage{placeins}
\usepackage{textcomp}
\usepackage{amsmath}
\usepackage{graphicx}
\usepackage{url}
\usepackage{multicol, blindtext}
\usepackage{booktabs}
\usepackage[table]{xcolor}
\usepackage{xcolor}
\usepackage{lipsum}
\usepackage{amssymb}
\usepackage{lineno}
\usepackage{algpseudocode}
\usepackage{amsfonts}
\usepackage{siunitx}
\usepackage{amsmath}
\usepackage{multirow}
\usepackage{wrapfig}
\usepackage{array}
\usepackage{color}
\usepackage{silence}
\usepackage[table]{xcolor}
\definecolor{lightgray}{gray}{0.9}
\usepackage{svg}

\begin{document}

\title{An Efficient Unsupervised Federated Learning Approach for Anomaly Detection in Heterogeneous IoT Networks}

\author{Mohsen Tajgardan, Atena Shiranzaei, Mahdi Rabbani, Reza Khoshkangini, Mahtab Jamali
\thanks{M. Tajgardan is with the Faculty of Electrical and Computer Engineering, Qom University of Technology, Qom, Iran (e-mail: tajgardan.m@qut.ac.ir). A. Shiranzaei is with the Department of Computer Engineering, Faculty of Industry and Mining (Khash), Sistan and Baluchestan University, Zahedan, Iran (e-mail: ashiranzaei@eng.usb.ac.ir). M. Rabbani is with the Canadian Institute for Cybersecurity, University of New Brunswick, NB, Canada (e-mail: m.rabbani@unb.ca). Reza Khoshkangini and Mahtab Jamali are with the Sustainable Digitalisation Research Centre (SDRC), Department of Computer Science and Media Technology, Faculty of Technology and Society, Malmo University, Sweden (emails: reza.khoshkangini, mahtab.jamali@mau.se).}}
\markboth{}
{Shell \MakeLowercase{\textit{et al.}}: A Sample Article Using IEEEtran.cls for IEEE Journals}


\maketitle

\begin{abstract}
\input{abstract}
\end{abstract}


\section{Introduction}	
\input{Introduction}

\section{Related Work}
\input{related_work}

\section{Data Presentation}
\input{data_representation}
    
\section{Proposed work}
\input{proposed_approach}

\section{Experimental Setup and Evaluation}
    	
\input{Evaluation}
\subsection{Model Explainability (SHAP results)}
\input{Explainability}

\section{Discussion and Conclusion}
\input{conclusion}


\bibliographystyle{elsarticle-num} 
\bibliography{ref}






\end{document}

%% file: abstract.tex
\textcolor{black}{Federated learning (FL) is an effective paradigm for distributed environments such as the Internet of Things (IoT), where data from diverse devices with varying functionalities remains localized while contributing to a shared global model. By eliminating the need to transmit raw data, FL inherently preserves privacy. However, the heterogeneous nature of IoT data, stemming from differences in device capabilities, data formats, and communication constraints, poses significant challenges to maintaining both global model performance and privacy. In the context of IoT-based anomaly detection, unsupervised FL offers a promising means to identify abnormal behavior without centralized data aggregation. Nevertheless, feature heterogeneity across devices complicates model training and optimization, hindering effective implementation.}
\textcolor{black}{In this study we propose an efficient unsupervised FL framework that enhances anomaly detection by leveraging shared features from two distinct IoT datasets: one focused on anomaly detection and the other on device identification, while preserving dataset-specific features. To improve transparency and interpretability, we employ explainable AI techniques, such as SHAP, to identify key features influencing local model decisions.}
\textcolor{black}{Experiments conducted on real-world IoT datasets demonstrate that the proposed method significantly outperforms conventional FL approaches in anomaly detection accuracy. This work underscores the potential of using shared features from complementary datasets to optimize unsupervised federated learning and achieve superior anomaly detection results in decentralized IoT environments.}

%% file: Introduction.tex
\label{sec:introduction}


\textcolor{black}{The rapid proliferation of Internet of Things (IoT) devices has transformed numerous sectors, including smart homes, healthcare, industrial automation, and critical infrastructure \cite{li2015internet, jamali2025context}. These devices originate from diverse environments, where they were developed by different vendors, and serve various functions, resulting in a highly heterogeneous ecosystem \cite{javed2020scalable, rabbani2024device}. Such heterogeneity introduces significant challenges in data processing, model training, and interoperability \cite{mahadik2024heterogeneous}. Furthermore, many IoT devices, such as medical sensors and industrial monitoring systems \cite{madhavan2025object, jamali2024video}, generate sensitive data, making the transfer of raw information to centralized servers undesirable due to privacy and security concerns.}

Federated Learning (FL) has emerged as a promising paradigm for addressing these challenges by enabling decentralized model training while keeping data localized. This approach is valuable in several IoT-based applications, particularly IoT anomaly detection, where identifying security threats in real time is essential without compromising data privacy \cite{mothukuri2021federated}.




\textcolor{black}{Despite its advantages, applying FL in IoT environments presents several challenges, primarily due to the inherent heterogeneity of IoT data. Devices vary in computational resources, communication bandwidth, and data formats, resulting in non-IID (non-independent and identically distributed) feature distributions between clients \cite{zhang2021client, rabbani2025lightweight}. This feature heterogeneity complicates model training and optimization, making it difficult to balance global model convergence with preservation of privacy. The challenge is further exacerbated in unsupervised anomaly detection, where the absence of labeled data requires the model to identify abnormal patterns without prior knowledge of attack behaviors.}

\textcolor{black}{Unsupervised anomaly detection in IoT networks is challenging due to their dynamic nature and the scarcity of labeled data \cite{lyu2020robustness}. Techniques such as clustering \cite{xu2005survey, khoshkangini2024hierarchical}, autoencoders \cite{hinton2006reducing}, and deep learning methods have proven effective in detecting new anomalies within federated learning. Nonetheless, these methods often face issues such as model heterogeneity, communication inefficiency, and convergence instability \cite{yang2019federated}. Feature-level heterogeneity, where devices generate data with differing types, distributions, and sampling rates can further degrade global model performance \cite{geyer2017differentially}. }


\textcolor{black}{While deep learning approaches such as heterogeneous neural networks, can integrate diverse architectures to handle feature variations to improve anomaly detection performance in decentralized IoT environments, most existing FL frameworks rely on homogeneous networks that require uniform input and output features across clients. This constraint often results in the loss and alteration of important shared feature characteristics, thereby limiting anomaly detection effectiveness. In contrast, handling heterogeneous data environments, where data types vary widely, often leads to inconsistencies that degrade global model performance.}

\textcolor{black}{To address these issues, we propose an unsupervised FL-based approach that improves performance anomaly detection by leveraging shared features from two distinct types of IoT datasets; The first one designed for detecting anomalous behaviors and the second for device identification, while preserving differences in input features. Instead of removing unique features, the proposed method integrates shared representations between clients to refine model weight optimization. This design mitigates the impact of feature heterogeneity while improving anomaly detection accuracy, and maintaining privacy.}

Our proposed system consists of four interconnected phases: (1) \textit{Semantic Data Refinement}, which transforms heterogeneous raw data into structured, learning-ready representations; (2) \textit{Federated Knowledge Aggregation, enabling decentralized clients to collaboratively construct a shared intelligence through privacy-preserving model fusion};
(3) \textit{Intelligent Device and Anomaly Profiling}, where learned representations are exploited to detect irregular behaviors and emerging threats; and
(4) \textit{Explainable Intelligence Assessment}, applying SHAP to reveal the key factors underlying model decisions and system behavior. The following research questions (RQs) define the core investigative objectives of the proposed approach:

\begin{itemize}

\item \textbf{RQ1: Federated learning under feature heterogeneity:}
 To what extent can heterogeneous and partially overlapping feature spaces be effectively integrated within a federated learning framework to construct an accurate and robust global model for device behavior modeling?

\item  \textbf{ RQ2: Detection effectiveness and explainability of federated models:}  
How effectively can the learned global federated model detect device-level anomalies, and to what extent can SHAP-based explainability provide meaningful insights into the key features driving these detection decisions?

\end{itemize}

Integrating datasets with diverse and partially overlapping feature spaces in a distributed environment, while maintaining high predictive accuracy, remains a key challenge in federated learning. To address this, we propose a federated learning framework capable of handling both homogeneous and heterogeneous clients, where datasets may differ in feature dimensionality and output classes. In the proposed framework, each client independently trains a local model on its respective dataset, and a central server aggregates the locally learned model parameters to construct a global model without requiring raw data exchange.

To mitigate feature heterogeneity, a dynamic weight adjustment mechanism is introduced to align model parameters across clients with differing feature dimensions. This alignment enables effective knowledge transfer while preserving the distinctive characteristics of each local dataset. The resulting global model is subsequently redistributed to clients and evaluated on local test data, demonstrating its ability to generalize across diverse data distributions.

Furthermore, the framework exploits shared and overlapping features among datasets to enhance collaborative learning. By leveraging common feature representations and complementary knowledge from both homogeneous and heterogeneous clients, the global model achieves improved accuracy and robustness. The weight adjustment process ensures compatibility across datasets with varying feature spaces, allowing shared features to reinforce learning while accommodating dataset-specific variations. This collaborative strategy illustrates how federated integration of diverse datasets can mutually improve model performance in distributed systems.

The main contributions of this paper are summarized as follows:

\begin{itemize}
\item \textbf{A unified federated learning framework for heterogeneous feature spaces:}
We propose a federated learning framework that seamlessly integrates both homogeneous and heterogeneous clients by employing dynamic weight adjustment mechanisms, enabling accurate global model construction across datasets with varying feature dimensions and output classes.

\item \textbf{Collaborative feature sharing with enhanced detection and interpretability:}  
 We introduce a shared-feature integration strategy that exploits overlapping feature representations to improve anomaly detection performance, complemented by SHAP-based explainability to provide transparent and interpretable insights into the global model’s decision-making process.
\end{itemize}

The rest of the paper is organized as follows: In Section \ref{sec:related} we review related works. We provide the dataset details in Section \ref{sec:data}. Section \ref{sec:proposedapproach} describes the proposed approach. In Section \ref{sec:results}, the experimental evaluation and results of our proposed approach are presented, followed up with a discussion and conclusion in Section \ref{sec:summary}.

%% file: related_work.tex
\label{sec:related}

Recent research on anomaly detection in IoT networks has explored a range of methodologies, including federated learning frameworks, unsupervised learning techniques, and deep learning approaches, to address challenges such as privacy preservation, data heterogeneity, and model performance. This section categorizes and reviews existing literature, highlighting key advances and identifying gaps that motivate our work.

\subsection{Federated Learning for Anomaly Detection}

Federated learning has emerged as a promising approach for anomaly detection in distributed environments, where data privacy and security are critical. In \cite{r1}, the authors introduce Fed-ANIDS, a federated learning-based framework that integrates anomaly detection using three types of autoencoders: simple, variational, and adversarial. This approach is designed to enhance intrusion detection in distributed networks while maintaining data privacy. Fed-ANIDS was evaluated on three datasets, USTC-TFC2016, CIC-IDS2017, and CSE-CIC-IDS2018, demonstrating high detection accuracy and low false alarm rates, and outperforming other GAN-based models. Shen et al. \cite{shen2024effective} propose FLEKD, an ensemble knowledge distillation-based FL framework for heterogeneous IoT networks, which outperforms conventional FL methods on CICIDS2019 by flexibly aggregating knowledge from diverse clients. Nguyen and Beuran \cite{nguyen2024fedmse} introduce FedMSE, a semi-supervised FL approach using a Shrink Autoencoder and Centroid classifier with MSE-based aggregation. This method significantly improves detection accuracy and efficiency on N‑BaIoT by prioritizing more accurate local models during aggregation. In \cite{r4}, a VHetNet-enabled asynchronous FL framework for UAV selection is proposed, enabling decentralized UAVs to collaboratively train a global anomaly detection model using local IoT data. The framework aims to reduce the execution time and computational overhead of federated learning while maintaining high detection accuracy and low energy consumption, as shown in evaluations on real-world datasets. The authors in \cite{r6} present an FL-based framework for IoT malware detection, supporting both supervised and unsupervised learning. Evaluated on the N-BaIoT dataset, their results demonstrate that diverse and large datasets can enhance model performance in both centralized and federated settings without compromising privacy.


\subsection{Techniques for Improving Federated Learning}

Several studies have proposed methods to enhance the performance of federated learning models by addressing class imbalance, data heterogeneity, and resource constraints. Weinger et al. \cite{r3} apply data augmentation to mitigate performance degradation in federated learning caused by class imbalance and device heterogeneity. Their study shows that data augmentation can significantly improve anomaly detection performance in IoT networks, as demonstrated using the TON-IoT and DS2OS datasets. Vahidian et al. \cite{r10} introduce a novel approach to handling data heterogeneity in federated learning by focusing on Non-IID (Non-Independent and Identically Distributed) partitioning. They argue that data heterogeneity can be beneficial for learning and propose a method that uses the Non-IID label skew to improve learning outcomes, challenging common assumptions in the federated learning community. Ren et al. \cite{r8} develop LVA-SP, a lightweight unsupervised intrusion detection model, based on a variational autoencoder. The model is designed for use in resource-constrained environments such as industrial control systems (ICSs), balancing detection accuracy with system resource usage. The LVA-SP model achieved high performance in terms of F1 score, time efficiency, and memory usage when tested on an ICS dataset.

\subsection{Unsupervised Techniques for Anomaly Detection}

Unsupervised learning is crucial for anomaly detection tasks, particularly in scenarios where labeled data is scarce or unavailable. In \cite{r2}, the authors present a federated learning framework that leverages unsupervised device clustering to train anomaly detection models for heterogeneous IoT networks. This clustering mechanism addresses heterogeneity while reducing communication overhead, and experiments show strong detection accuracy with a low false positive rate across multiple attack scenarios. The study in \cite{r7} employs basic autoencoders (bAEs) for dimensionality reduction and a three-stage detection pipeline using one-class SVM, deep autoencoder, and DBSCAN clustering. This approach was evaluated in the CIC-IDS2017 and CSE-CIC-IDS2018 datasets, demonstrating its effectiveness in identifying complex attack patterns.

\subsection{Deep Learning Approaches for Anomaly Detection}

Deep learning techniques have shown great promise in detecting complex anomalies in IoT data due to their ability to learn intricate patterns. Yaras and Dener \cite{yaras2024iot} propose a hybrid CNN–LSTM model for DDoS attack detection. The model was trained and evaluated on the CICIoT2023 and TON\_IoT datasets, achieving high detection rates for both attack identification and type classification, demonstrating robust performance across multiple dataset scenarios. Wang et al. \cite{r5} present an FL-based anomaly detection approach using deep neural networks (DNNs), where only model weights are shared with the central server. This preserves privacy while outperforming traditional deep learning in accuracy and false alarm reduction, evaluated on the IoTBotnet 2020 dataset.

\subsection{Integration of Multiple Data Sources and Feature Engineering
Combining}
Combining data from multiple sources with advanced feature engineering can improve model robustness and accuracy. Studies such as \cite{carter2022fusion} and \cite{davis2021ensemble} demonstrate that data fusion and ensemble learning can enhance anomaly detection performance. These techniques aggregate information from diverse datasets to create more robust models. However, integrating heterogeneous feature spaces remains challenging. Recent approaches focus on leveraging shared features across datasets while preserving individual characteristics to enhance model generalization \cite{adams2023features}.


While FL-based methods have advanced IoT anomaly detection, most either assume homogeneous feature spaces or handle heterogeneity by transforming, clustering, or discarding unique dataset-specific features. Such approaches often lead to the loss of critical information and limit the model’s ability to capture nuanced anomalies. Moreover, methods like FLEKD and FedMSE are typically designed for supervised or semi-supervised settings, relying on labeled data for effective training. Fully unsupervised FL approaches for IoT anomaly detection remain scarce, and existing ones rarely focus on leveraging shared features from multiple heterogeneous datasets. In contrast, our method operates entirely in an unsupervised manner, directly integrates datasets with both shared and unique features, dynamically aligns heterogeneous feature dimensions through weight adjustment, and preserves dataset-specific attributes. This design enables the global model to exploit complementary strengths from different clients without sacrificing accuracy or privacy, a capability largely unexplored in prior studies.

%% file: data_representation.tex
\label{sec:data}
This study evaluates the proposed anomaly detection method using three publicly available IoT network intrusion datasets:

  \begin{itemize}
            \item {CICIoT2022:}
            This dataset contains 48 features that identify approximately 28 devices. For this work, only the 12 devices with the highest number of samples are retained, and non-informative records are excluded.

            \item {CICIoT2023:}
            This dataset was collected from a network of 150 devices executing 33 attack types across 7 categories. In this paper, all attacks, regardless of their type, are grouped into one class, and normal traffic is placed into another class. Each traffic sample consists of 46 features that serve as input for the autoencoder.

            \item {CICIoT-DIAD 2024:}
            In this dataset, a structure file and an HTTP Flood attack file from the DDoS category are used, which contain 135 features. The first 57 features are discarded and the final 78 features are used for anomaly detection.
  \end{itemize}

In datasets, benign traffic is labeled as 0 and attack traffic as 1.
Table \ref{table:dataset} summarizes the number of features and samples for each class in the datasets. The common features among the three datasets (highlighted in blue) are used as the shared feature space in the federated learning process.


Although the datasets differ in feature naming, number of samples, and year of collection, they share a similar nature, each is derived from network traffic generated by IoT devices. Our results indicate that merging such datasets, despite structural differences, can yield robust performance in federated anomaly detection, suggesting that combining datasets of similar origin can enhance the generalization capability of FL-based models.


    \begin{table*}
        \caption{Details of the dataset.}
        \label{table:dataset}
        \renewcommand{\arraystretch}{1.3}
        \scriptsize
        \setlength{\tabcolsep}{3pt} 
        \centering
        \begin{tabularx}{\linewidth}{
            >{\centering\arraybackslash}X
            >{\centering\arraybackslash}p{0.10\linewidth}
            >{\centering\arraybackslash}p{0.10\linewidth}
            >{\centering\arraybackslash}p{0.12\linewidth}
            >{\centering\arraybackslash}p{0.12\linewidth}
            >{\centering\arraybackslash}p{0.12\linewidth}
            >{\centering\arraybackslash}p{0.12\linewidth}
        }
            \hline
            \textbf{Datasets} & \textbf{No. Features} & \textbf{Class Name} & \textbf{No. Samples For All} & \textbf{No. Samples Train} & \textbf{No. Samples Test}& \textbf{No. Samples validation}\\
            \hline
            CICIoT-DIAD 2024~\cite{rabbani2024device} & 78 & Benign & 297063&232904 & 6000 & 58226\\
            (Anomaly Detection) & & Attack & 290346 &227518 & 6000 &56880\\
            Heterogeneous Client &  &  &  & &  &\\
            \hline
            CICIOT Dataset 2023~\cite{neto2023ciciot2023} & 46 & Benign & 919206&730577 & 6000 & 182645\\
            (Anomaly Detection) & & Attack & 152686 &117444 & 6000 &29361\\
            Heterogeneous Client &  &  &  & &  &\\
            \hline 
             &  & simcam & 34217 &26584&1000 &6646\\
             &  & homeeyecam & 27619&21310&1000  &5328\\
             &  & arloqcam & 21948&16779&1000  &4195\\
             &  & arlobasecam & 20520&15634&1000  &3909\\
             &  & luohecam & 20076&15275&1000  &3819\\
            CICIOT Dataset 2022~\cite{dadkhah2022towards} & 48 & amcrest & 18362&13912&1000  &3478\\
            (Device Identification) &  & dlinkcam & 14099&10508&1000  &2627\\
             Homogeneous Client &  & heimvisioncam & 13787&10257&1000  &2564\\
             &  & eufyhomebase & 11007&8042&1000  &2010\\
             &  & netatmocam & 10625&7745&1000  &1936\\
             &  & nestcam & 7490&5242&1000  &1311\\
            \hline
        \end{tabularx}
        
    \end{table*}


\begin{table*}
    \caption{Comparison of features across three datasets, features in blue represent common attributes.}
    \label{table:dataset}
    \centering
    \begin{minipage}{0.48\textwidth}
        \centering
        \textbf{(a) CICIoT2022 Device Identification Dataset  ~\cite{dadkhah2022towards}}
        \BlankLine
        \renewcommand{\arraystretch}{1.1}
        \begin{tabular}{|p{0.3cm}|p{3cm}|p{0.3cm}|p{3cm}|}
            \hline
            \textbf{No.} & \textbf{Feature Name} & \textbf{No.} &  \textbf{Feature Name}\\
            \hline
            1 & \textcolor{blue}{total\_length} & 26 & max\_et \\
            2 & L4\_udp & 27 & med\_et \\
            3 & L7\_http & 28 & average\_et \\
            4 & L7\_https & 29 & skew\_et \\
            5 & port\_class\_src & 30 & kurt\_et \\
            6 & port\_class\_dst & 31 & \textcolor{blue}{var} \\
            7 & pck\_size & 32 & q3 \\
            8 & ip\_dst\_new & 33 & \textcolor{blue}{q1} \\
            9 & ethernet\_frame\_size & 34 & iqr \\
            10 & ttl & 35 & sum\_e \\
            11 & L4\_tcp & 36 & \textcolor{blue}{min\_e} \\
            12 & \textcolor{blue}{protocol} & 37 & \textcolor{blue}{max\_e} \\
            13 & source\_port & 38 & med \\
            14 & dest\_port & 39 & \textcolor{blue}{average} \\
            15 & \textcolor{blue}{DNS\_count} & 40 & \textcolor{blue}{inter\_arrival\_time} \\
             16 & NTP\_count &  41 & kurt\_e\\
             17 & \textcolor{blue}{ARP\_count} & 42 & var\_e\\
             18 & \textcolor{blue}{cnt} & 43 & q3\_e\\
             19 & L3\_ip\_dst\_count & 44 & \textcolor{blue}{q1\_e}\\
             20 & most\_freq\_d\_ip & 45 & iqr\_e\\
             21 & most\_freq\_prot & 46 & epoch\_timestamp\\
             22 & most\_freq\_sport & 47 & skew\_e\\
             23 & most\_freq\_dport & 48 & time\_since\_previously\_ displayed\_frame\\
             24 & \textcolor{blue}{sum\_et} & 49 & target(device identification)\\
             25 & min\_et &  & \\
            
            \hline
        \end{tabular}
    \end{minipage}
    \hfill
    \begin{minipage}{0.48\textwidth}
        \centering
        \textbf{(b) CICIoT2023 Attack Dataset ~\cite{neto2023ciciot2023}}
        \BlankLine
        \renewcommand{\arraystretch}{1.1}
        \begin{tabular}{|p{0.3cm}|p{3cm}|p{0.3cm}|p{3cm}|}
            \hline
            \textbf{No.} & \textbf{Feature Name} & \textbf{No.} &  \textbf{Feature Name}\\
            \hline
            1 & \textcolor{blue}{flow\_duration} & 26 & IRC \\
            2 & Header\_Length & 27 & TCP \\
            3 & Duration & 28 & UDP \\
            4 & Rate & 29 & DHCP \\
            5 & Srate & 30 & fin\_count \\
            6 & Drate & 31 & \textcolor{blue}{Variance} \\
            7 & fin\_flag\_number & 32& IPv \\
            8 & syn\_flag\_number & 33& LLC \\
            9 & rst\_flag\_number & 34& SMTP \\
            10 & psh\_flag\_number & 35& Std \\
            11 & ack\_flag\_number & 36& \textcolor{blue}{Min} \\
            12 & \textcolor{blue}{Protocol\_type} & 37& \textcolor{blue}{Max} \\
            13 & ece\_flag\_number & 38& Tot size \\
            14 & cwr\_flag\_number & 39& \textcolor{blue}{AVG} \\
            15 & \textcolor{blue}{DNS} & 40& \textcolor{blue}{IAT} \\
            16 & syn\_count & 41 & urg\_count\\
             17 & \textcolor{blue}{ARP} & 42 & Magnitue\\
             18 & \textcolor{blue}{Number} & 43 & Covariance\\
             19 & rst\_count & 44 & ICMP\\
             20 & HTTP & 45 & Weight\\
             21 & HTTPS & 46 & Radius\\
             22 & ack\_count & 47 & Target(anomaly detection)\\
             23 & Telnet &  & \\
             24 & \textcolor{blue}{Tot sum} &  & \\
             25 & SSH &  & \\
            \hline
        \end{tabular}
    \end{minipage}
    \hfill
    \vspace{4mm}
    \begin{minipage}{0.99\textwidth}    
        \centering
        \textbf{(c) CIC IoT-DIAD 2024 Attack Dataset~\cite{rabbani2024device}}
        \BlankLine
        \renewcommand{\arraystretch}{1.1}
        \begin{tabular}{|p{0.3cm}|p{3cm}|p{0.3cm}|p{3cm}|p{0.3cm}|p{3cm}|p{0.3cm}|p{3cm}|}
            \hline
            \textbf{No.} & \textbf{Feature Name} & \textbf{No.} &  \textbf{Feature Name} & \textbf{No.} & \textbf{Feature Name} & \textbf{No.} &  \textbf{Feature Name}\\
            \hline
            1 & stream\_jitter\_1\_var & 26 & channel\_10\_count& 51 & src\_ip\_60\_mean	& 76 &  q1\_p	\\
            2 & stream\_5\_count & 27 & channel\_10\_mean	 & 52 & src\_ip\_60\_var	& 77 &  iqr\_p	\\
            3 & stream\_5\_mean & 28 & channel\_10\_var	& 53 & src\_ip\_mac\_60\_count	& 78 &  l3\_ip\_dst\_count\\
            4 & stream\_5\_var & 29 & stream\_jitter\_10\_sum	& 54 & src\_ip\_mac\_60\_mean	& 79 & Target(anomaly detection) \\
            5 & src\_ip\_5\_count & 30 & stream\_jitter\_10\_mean	& 55 & src\_ip\_mac\_60\_var	&  &  \\
            6 & src\_ip\_5\_mean & 31 & stream\_jitter\_10\_var	& 56 & channel\_60\_count	&  &  \\
            7 & src\_ip\_5\_var & 32 & stream\_30\_count	& 57 & channel\_60\_mean	&  &  \\
            8 & src\_ip\_mac\_5\_count & 33 & stream\_30\_mean	& 58 & channel\_60\_var	&  &  \\
            9 & src\_ip\_mac\_5\_mean & 34 & stream\_30\_var	 & 59 & stream\_jitter\_60\_sum	 &  &  \\
            10 & src\_ip\_mac\_5\_var & 35 & src\_ip\_30\_count	& 60 & stream\_jitter\_60\_mean	&  &  \\
            11 & channel\_5\_count & 36 & src\_ip\_30\_mean	& 61 & stream\_jitter\_60\_var	&  &  \\
            12 & channel\_5\_mean & 37 & src\_ip\_30\_var	& 62 & ntp\_interval	&  &  \\
            13 & channel\_5\_var & 38 & src\_ip\_mac\_30\_count	& 63 & most\_freq\_spot	&  &  \\
            14 & stream\_jitter\_5\_sum & 39 & src\_ip\_mac\_30\_mean	& 64 & min\_et	&  &  \\
            15 & stream\_jitter\_5\_mean & 40 & src\_ip\_mac\_30\_var	& 65 & \textcolor{blue}{q1}	&  &  \\
            16 & stream\_jitter\_5\_var & 41 & channel\_30\_count	& 66 & min\_e	&  &  \\
            17 & stream\_10\_count & 42 & channel\_30\_mean	& 67 & var\_e	&  &  \\
            18 & stream\_10\_mean & 43 & channel\_30\_var	& 68 & \textcolor{blue}{q1\_e}	&  &  \\
            19 & stream\_10\_var & 44 & stream\_jitter\_30\_sum	& 69 & sum\_p	&  &  \\
            20 & src\_ip\_10\_count & 45 & stream\_jitter\_30\_mean	& 70 & min\_p	&  &  \\
            21 & src\_ip\_10\_mean & 46 & stream\_jitter\_30\_var	& 71 & max\_p	&  &  \\
            22 & src\_ip\_10\_var & 47 & stream\_60\_count& 72 & 	med\_p	&  &  \\
            23 & src\_ip\_mac\_10\_count & 48 & stream\_60\_mean	& 73 & average\_p	&  &  \\
            24 & src\_ip\_mac\_10\_mean & 49 & stream\_60\_var	& 74 & var\_p	&  &  \\
            25 & src\_ip\_mac\_10\_var & 50 & src\_ip\_60\_count	& 75 & 	q3\_p	&  &  \\
            
            \hline
        \end{tabular}
    \end{minipage}
\end{table*}

%% file: proposed_approach.tex
\label{sec:proposedapproach}

\begin{figure}
    \centering
    \includegraphics[width=1\linewidth]{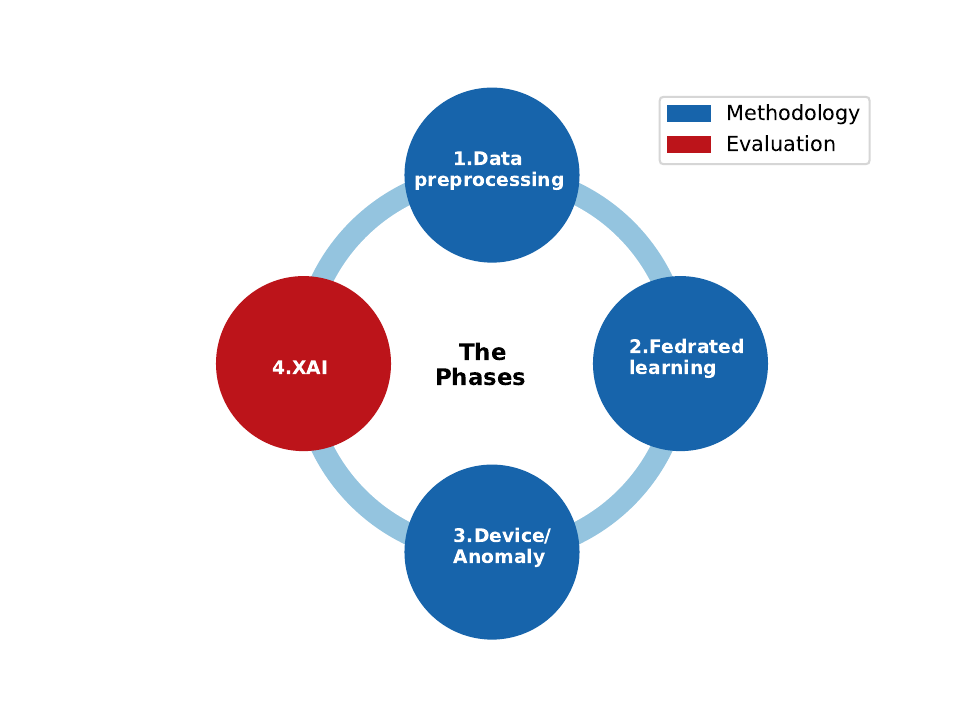}
    \caption{The high-level life cycle of the proposed approach.}
    \label{fig:phases}
\end{figure}

As illustrated in Figure~\ref{fig:phases}, the proposed approach is structured into four main phases: Semantic Data Refinement, Federated Knowledge Aggregation, Intelligent Device and Anomaly Profiling, and Explainable Intelligence Assessment. The final phase is integrated into the evaluation process to enhance the interpretability and trustworthiness of the model outcomes. The details of each phase are discussed in the following sections.

\subsection{Data Preprocessing}

Data preprocessing transforms raw data into a clean, structured format suitable for model training. In this study, three datasets are deployed across different clients: CICIoT2022 for device identification, CICIoT2023 and CICIoT-DIAD 2024 for anomaly detection.

For each client:

\begin{itemize}
    \item \textbf{Feature–Label Separation and Normalization}: The dataset is loaded, features are separated from labels, and feature values are normalized using Min-Max scaling.
    \item \textbf{Balancing the Datasets}: For anomaly detection, we select 6,000 normal and 6,000 attack samples. For device identification, we retain the 11 most represented devices and extract 1,000 samples from each to ensure balanced representation in testing.
\end{itemize}

\subsection{Federated Learning}
\begin{figure}
    \centering
    \includegraphics[width=1\linewidth]{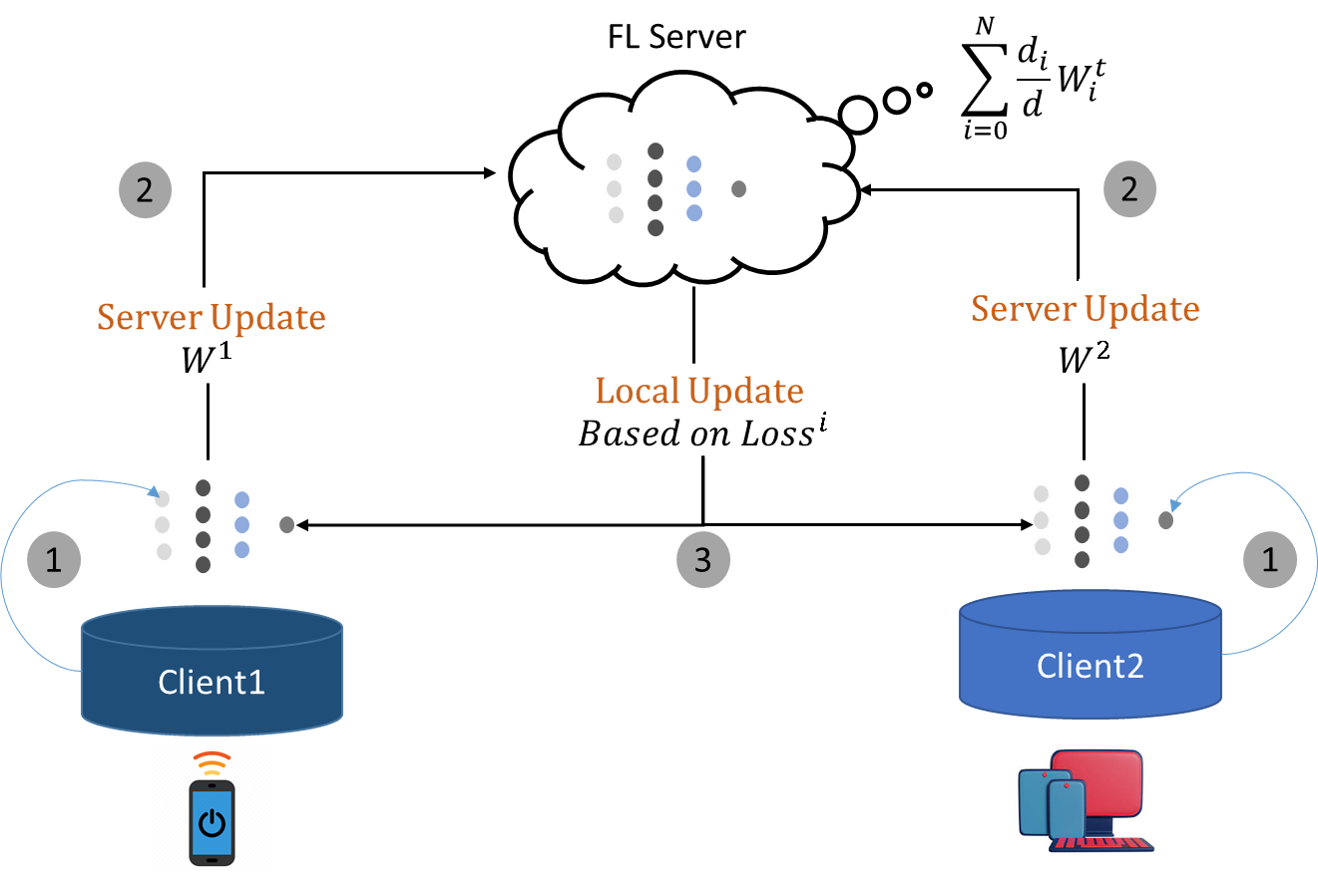}
    \caption{A general overview of the federated learning system}
    \label{fig:FL}
\end{figure}

In this phase, the clients independently and in parallel train their local models. Then the trained model weights are sent to the server and aggregated. The global model is constructed and distributed to the clients to evaluate their local test data. Figure \ref{fig:FL}, illustrates a general overview of the federated learning system. Client 2, uses the CICIoT2023 dataset for anomaly detection. The input is the same as the server, with 46 features, and the output is a binary value (0 or 1), representing normal and attack traffic in the IoT. Client 1, on the other hand, has different features, with 48 inputs and 11 outputs, identifying different types of device traffic, and the next client is not shown in the image because of its similarity to client 2.


\label{sec:approach}


\begin{figure*}
    \centering
    \includegraphics[width=0.8\linewidth]{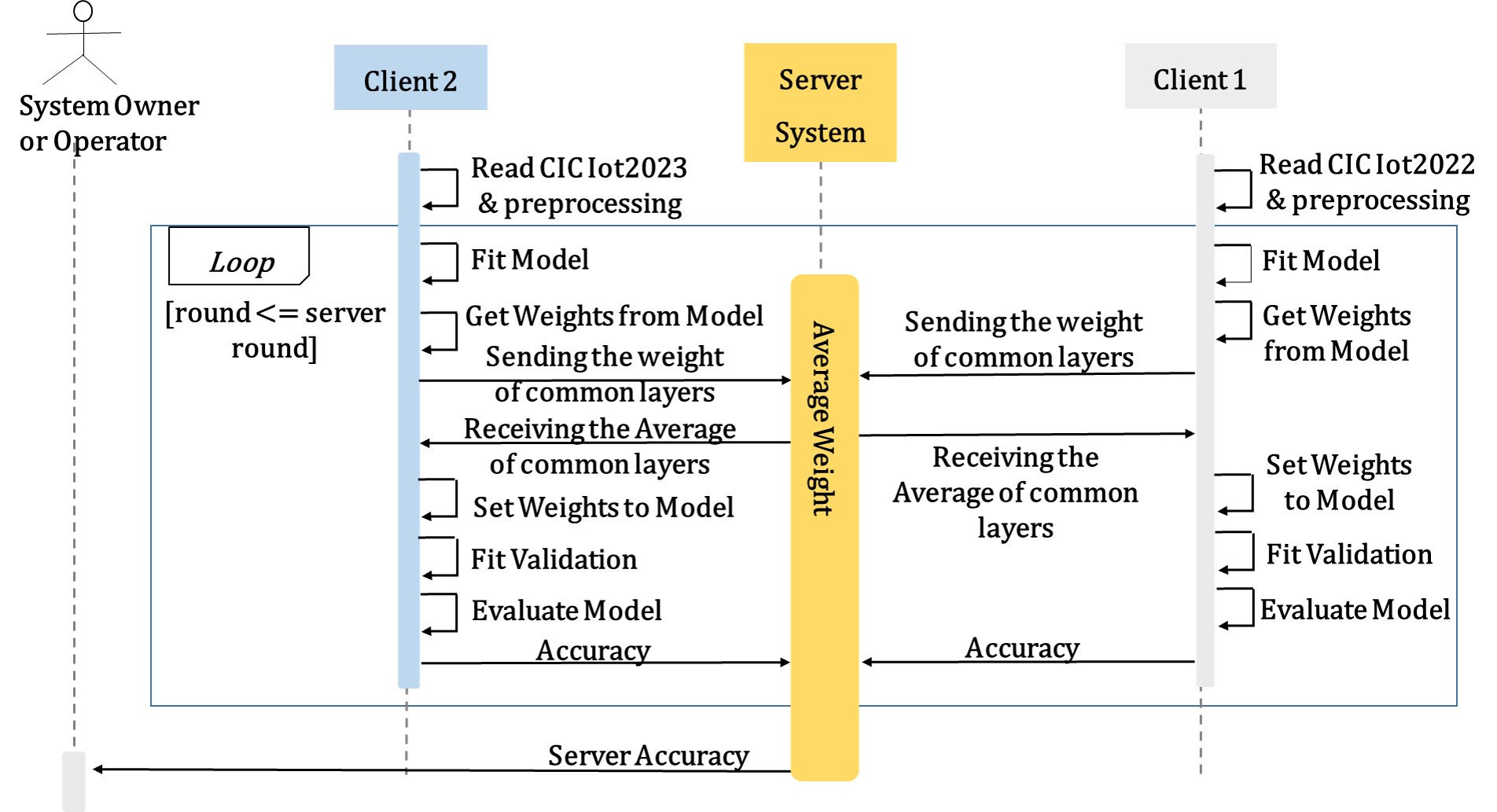}
    \caption{An external view of the federated learning framework.}
    \label{fig:Flow}
\end{figure*}

In the following, Figure \ref{fig:Flow} illustrates an external view of the federated learning system, showing the interactions between the actors involved in the execution of the program. Initially, the operator places the server, and the three clients are placed in standby mode by the operator. The server performs a handshake to detect the presence of clients and then waits for the client models to be trained and for the corresponding weights to be received, enabling the creation of the global model, which will be sent back to the clients. The server is unaware of the existence of heterogeneous clients, and the server talks to the clients over common layers. 


\subsubsection{\textbf{Local Model Training}}

\begin{figure}
    \centering
    \includegraphics[width=1\linewidth]{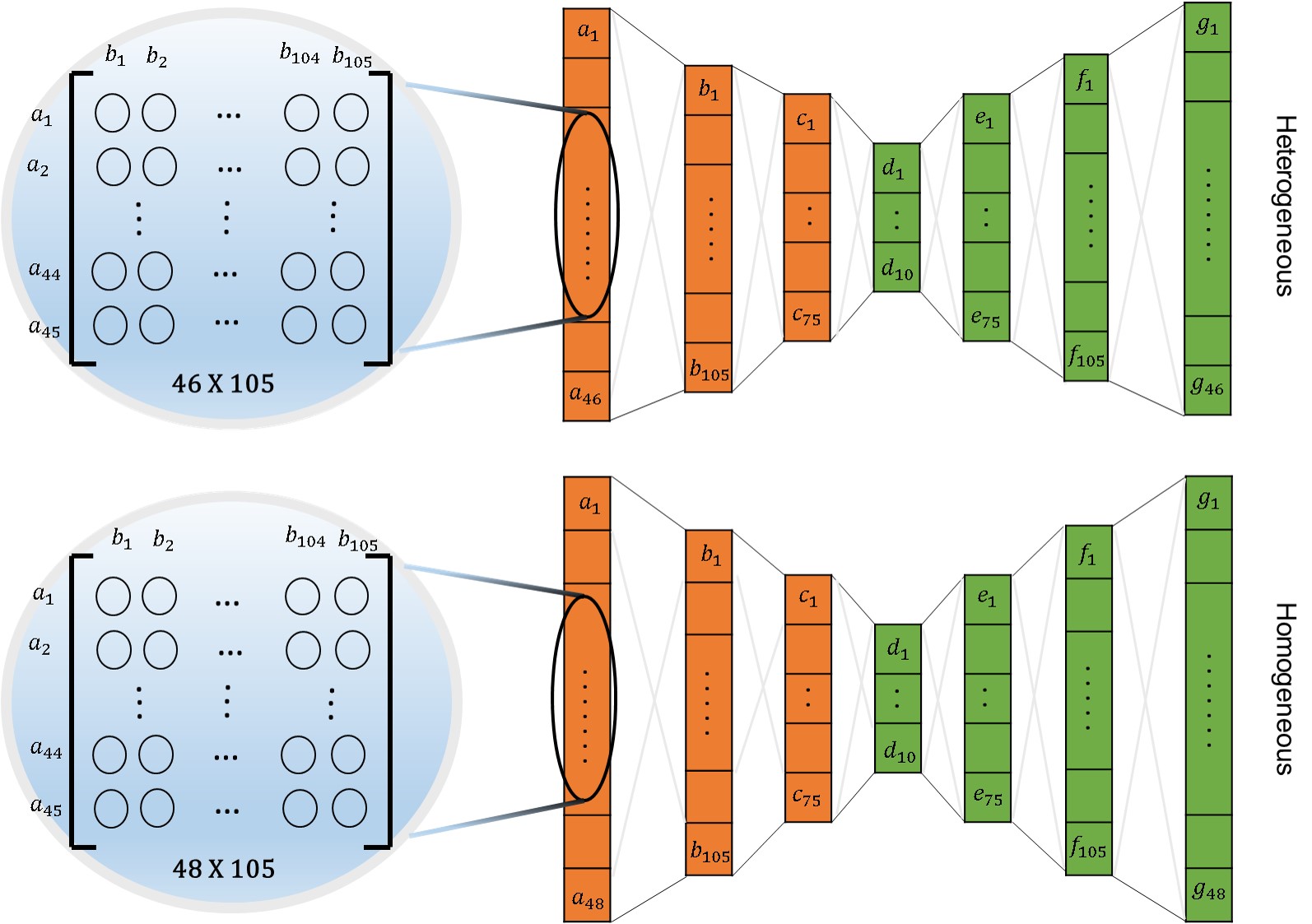}
    \caption{An internal view of the clients and how different weights have been set.}
    \label{fig:reshape}
\end{figure}

\begin{itemize}
    \item Each client employs an unsupervised autoencoder architecture (as shown in Figure~\ref{fig:reshape}), using Adam optimizer and mean squared error (MSE) loss:

    \begin{equation}
        \text{MSE} = \frac{1}{n} \sum_{i=1}^{n} (x_i - \hat{x}_i)^2
        \label{eq:mse}
    \end{equation}

    \item For heterogeneous models, it is enough to send only the weight of the matrix with common dimensions to the server, and the server returns their average after calculating them. The weight of the matrix of layers with common dimensions is placed next to the different layers and now it is the turn to equalize the weights of these two cases, which is solved by training with validation data. In our approach, client models are designed to be as structurally uniform as possible. However, differences in input feature counts lead to variations in the first and last layers, resulting in incompatible weight matrices when aggregating models on the server. Since the method is unsupervised, the output layer mirrors the input layer, and intermediate hidden layers remain identical across clients. The autoencoder architecture is symmetric, with layers gradually reducing to a bottleneck latent layer that captures compressed representations of the input. As shown in Figure~\ref{fig:reshape}, all layers are identical except for the first and last layers, which are structured based on the number of input features. This issue can be resolved on the heterogeneous client side by removing the different weight matrices, and only the average of the common layers is returned. And finally, for the test data, we receive it from the latent layer and inject it to K-means for clustering.

    In Figure \ref{fig:reshape}, the internal view of the clients and how the different weights are set in the clients are shown. In Client 1, which uses the CICIoT2022 dataset, the weights of the initial and final layers are different from the weights of the client (CICIoT2023). In the initial layer, a matrix of size 48×105 is formed, while the weight matrix of the initial layers of Client 2 is 46×105. In addition, there is a bias of 48, which is different from the bias of 46. In the final layer, the matrix is 105×48 with a bias of 48 and should be 105×46 and the bias of 46 remains in the clients. The rest of the weights and biases are transmitted to the server with equal dimensions. And the server also returns the mean of equal dimensions between all clients. The model is rebuilt in the clients and equal weights are replaced. The challenge is that the weights must be modified to produce the desired output, those weights must be trained with the validation data, and now the model is ready to decode the test data.

    In the first paragraph of Algorithm~\ref{alg:propose}, after models have been trained, the layers with common dimensions are averaged and placed between different layers because the weights are not consistent. The two stages are trained with the validation data, and now the model is ready to encode the test data by the latent layer. After encoding the data, it transferred to K-means for clustering. The output classification is optimized, and the metrics are calculated.

    In Table~\ref{table:paper_define}, we provide a comprehensive reference for the datasets, algorithmic parameters, and mathematical symbols used in this work, and in Table~\ref{table:paper_config}, the values of the parameters for this model are presented.


\end{itemize}



\begin{table}
    \centering
    \begin{tabular}{|l| m{6.2cm}|}
        \hline
        \textbf{Term} & \textbf{Definition} \\
        \hline
        Client 1 &  CICIoT2022 Device Identification Dataset~\cite{dadkhah2022towards}.\\
        Client 2 &  CICIoT2023 Attack Dataset ~\cite{neto2023ciciot2023}.\\
        
        Client 3 &  CIC IoT-DIAD 2024 Device Identification \& Attack Dataset~\cite{rabbani2024device}.\\
        Homogeneous & A client with the same layer dimensions as the server.\\ 
        Heterogeneous & A client with at least one layer dimension different from the server.\\ 
        Round & One iteration of local training followed by global aggregation.\\
        Epoch & One iteration of local client training.\\
        \hline
    \end{tabular}
    \caption{Key terminology used in this study.}
    \label{table:paper_define}
\end{table}

\begin{table}
    \centering
    \begin{tabular}{|l| m{5cm}|}
        \hline
        \textbf{Parameter} & \textbf{Value} \\
        \hline
        Optimizer &  Adam\\
        Loss function &  mean squared error (MSE)\\
        Server Rounds &  21\\
        Client Train Epochs &  2\\
        Client Validations Epochs &  2\\
        Client Train Size &   80\%\\
        Client Validation size &   20\%\\
        Client Test Size &   $\approx$12000\\
        Activation &  RelU\\
        Random State & 42\\
        Clients Number & 3\\
        Learning Rate & 1e-3\\
        Batch Size & 32 \\
        $d_{2022}$ & equal 1 in Equation ~\ref{eq:update_server}\\
        $d_{2023}$ & equal 98 in Equation ~\ref{eq:update_server}\\
        $d_{2024}$ & equal 1 in Equation ~\ref{eq:update_server}\\

        \hline
    \end{tabular}
    \caption{Configs used in our study.}
    \label{table:paper_config}
\end{table}

\begin{algorithm}
\caption{Proposed Federated Learning Algorithm}
\label{alg:propose}
\KwIn{Datasets}
\KwOut{Client and Server metrics}
\BlankLine

\For{each client $c \in clients$}{
    Train $c$\texttt{.model} with Train data from $c$\texttt{.dataset}\;
    $i \gets 1$\;
    \For{each layer $l$ in $c$\texttt{.model}}{
        \If{$l$ is not the first or last layer}{
            $w[i] \gets w[i] + l$\texttt{.get\_weight()}\;
            $i \gets i + 1$\;
        }
    }
}

\For{each client $c \in clients$}{
    $w[0] \gets c$\texttt{.layer.first.getWeight()}\;
    $w[i+1] \gets c$\texttt{.layer.last.getWeight()}\;
    $c$\texttt{.model.set\_weight($w$)}\;
    
    Repair the model weights by training with validation data from $c$\texttt{.dataset}\;
    Encode test data using $c$\texttt{.model} from $c$\texttt{.dataset}\;
    Cluster encoded test data using K-means\;
    Optimize clustering of encoded test data\;
    Calculate client metrics\;
}

Calculate server metrics\;

\KwRet{Client and Server metrics}

\end{algorithm}


\subsubsection{\textbf{Global Model Update}}
Typically, training is not performed on the server; instead, all training occurs on the clients. Once the clients have trained their models, the weights are transferred to the server, where the server averages the weights according to Equation ~\ref{eq:update_server}. The client weights are combined using a weighted sum, which in some FL approaches is based on the training loss. In these cases, the weight of each client's model is multiplied by its training loss divided by the total loss across all clients. In other FL approaches, the weights are adjusted based on the number of samples each client uses in training, specifically, the model weight of each client is multiplied by the ratio of its training samples to the total number of training samples across all clients.


\textit{Federating Averaging Algorithm:}
When the server receives all the weights of the client model, it is crucial to aggregate them based on their importance. This importance can be determined by either the model's training loss or by the number of samples used during training. The importance is quantified probabilistically. For example, according to Equation ~\ref{eq:update_server}, the importance of a client model is calculated as the ratio of its loss to the total loss in all client models.

\textcolor{black}{ The server aggregates only the parameters of the \emph{common layers} across clients using a sample-size–weighted average. Let \(W_c^{(t)}\) denote the vector of parameters of the common layers for client \(c\) after round \(t\), and let \(d_c\) be the impact of each client \(c\) in that round. the values are given in Table~\ref{table:paper_config}. The global parameters are:}

\begin{equation}
    \text{server\_weight}^{t+1} = \sum_{c=1}^{n} \left( \frac{d_c}{\sum_{i=1}^{n} d_i} \right) W_{c}^{t}
    \label{eq:update_server}
\end{equation}
\textcolor{black}{Client-specific layers (for example, input and output layers with different dimensions) are not aggregated and remain local.} 

Despite the differences between the databases, they share a significant number of similar features. These shared features, used through the federated learning framework, help improve the performance of global models, particularly in anomaly detection when tested against experimental data, producing better results than the initial models.

Finally, the model is re-trained using validation data, ensuring the weight adjustments are finalized and resulting in improved performance in the final process.


    \begin{table}
    \caption{Autoencoder Architecture Details for CICIOT2023 heterogeneous Client Anomaly Detection Dataset.}
        \label{table:summary_atoencoder}
    \renewcommand{\arraystretch}{1.3}
        
        \begin{tabular}{|p{3.5cm}|p{2.5cm}|p{1.5cm}|}
            \hline
            Layer (type) & Output Shape & Param \# \\
            \hline
            input\_layer InputLayer) & (None, 45)&0\\
            \hline
            First (Dense) & (None, 105)&4,830\\

            \hline
            dense (Dense) & (None, 90)&9,540\\

            \hline
            dense\_1 (Dense) & (None, 75)&6,825\\

            \hline
            dense\_2 (Dense) & (None, 60)&4,560\\

            \hline
            bottleneck (Dense) & (None, 10)&610\\
            \hline
            dense\_3 (Dense) & (None, 60)&660\\

            \hline
            dense\_4 (Dense) & (None, 75)&4,575\\

            \hline
            dense\_5 (Dense) & (None, 90)&6,840\\

            \hline
            dense\_6 (Dense) & (None, 105)&9,555\\

            \hline
            Last (Dense) & (None, 46)&4,876\\

            \hline

\end{tabular}
\begin{tablenotes}
      \scriptsize
\item   {Total params: 52,765 (206.11 KB)}
\item Trainable params: 52,765 (206.11 KB)
\item Non-trainable params: 0 (0.00 KB)
    \end{tablenotes}
\end{table}

\begin{figure}
    \centering
    \includegraphics[width=0.8\linewidth]{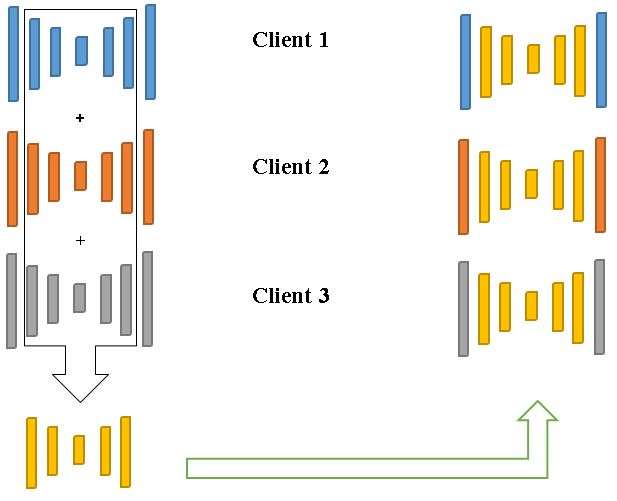} 
    \caption{Federated aggregation over common layers across clients.}
    \label{fig:fd_avg}
\end{figure}

As shown in Figure~\ref{fig:fd_avg}, in this federated learning phase, we average the weights of the layers that are similar and leave the dissimilar layers at the beginning and end untouched, and then we average the similar layers among the unchanged dissimilar layers and create a new weight layer for each client. We train the new autoencoder with a few steps with validation data to modify its weights and form a relationship between the weights.

\begin{figure}
    \centering
    \includegraphics[width=0.8\linewidth]{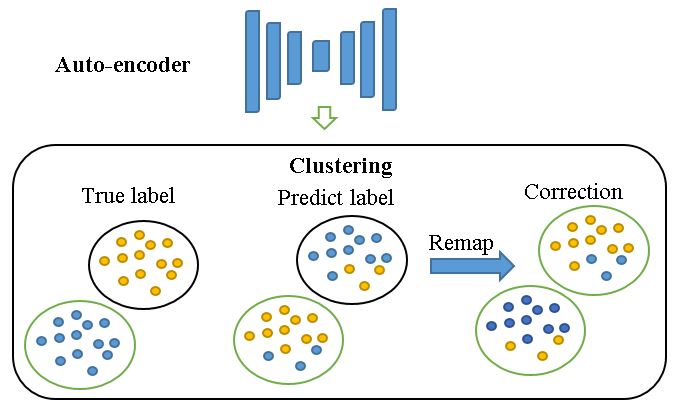}
    \caption{Overview of the clustering process on latent representation given by Auto-encoder.}
    \label{fig:fd_clustring}
\end{figure}

In Figure~\ref{fig:fd_clustring}, We use the trained autoencoder, which has been trained on the full dataset, and then pass the test data through it. From the middle layer of the decoder, we extract the test data representations and apply K-means clustering. At the bottom of the figure, you can see the K-means clustering results, which may assign labels differently in the detection simulation. If such a label mismatch occurs, we resolve it by swapping the labels.

\subsection{\textcolor{black}{Device Identification and Anomaly Detection}}
\textcolor{black}{After each communication round, every client rebuilds its local model by merging the server-averaged weights of the common layers with its client-specific input and output layers, followed by brief validation fine-tuning to align the adjusted weights. For inference, each client encodes test samples through the autoencoder and extracts the latent vector from the bottleneck layer. K-means clustering is then applied to these latent representations: k = 11 for device identification (CICIoT2022) and k = 2 for anomaly detection (CICIoT2023 and CICIoT-DIAD 2024).}

\textcolor{black}{\textbf{Evaluation sets:} For device identification (CICIoT2022), the evaluation set is a balanced sample of 11,000 records, obtained by randomly selecting 1,000 samples per device across 11 device classes. For anomaly detection (CICIoT2023 and CICIoT-DIAD 2024), the evaluation set contains 12,000 samples, evenly split between benign (6,000) and attack (6,000) traffic. This balanced sampling ensures a fair comparison between classes.}

\textbf{Label alignment:} Because K-means assigns arbitrary cluster indices, we apply label alignment prior to scoring. In the binary case (anomaly detection), we evaluate both the original predictions and their inverted mapping and retain the mapping with higher accuracy. In the multi-class case (device identification), we use a frequency-based mapping that selects the dominant predicted label per true-class block and resolves conflicts by choosing the next most frequent unused label. Formal definitions and pseudocode appear in Section V-A.

The aligned predictions are then used to compute accuracy, precision, recall, and F1 (see Section V-C), and results are reported per dataset using the best global round.


%% file: Evaluation.tex
\label{sec:results}

This section shows the experimental evaluation of the proposed approach over multiple steps as follows:



\subsection{Label Alignment for K-means Predictions}

During evaluation, encoded test data are fed to the K-means model for classification. Each test sample has a ground-truth label, while K-means assigns a predicted label. In unsupervised settings, the numerical values of predicted labels may not align with the true labels, even if the class groupings are semantically correct. This mismatch becomes increasingly complex as the number of classes grows, making manual label correction impractical and time-consuming.


\subsubsection{Anomaly Detection (Binary Case)}
In binary classification (e.g., normal = 0, anomalous = 1), K-means predictions may be numerically inverted with respect to the ground-truth labels. To detect and correct this, we compute the accuracy in two cases: the original label assignment (Equation 3) and the inverted assignment (Equation 4). The label mapping that yields the higher accuracy is selected as the correct one, as defined in Equation 5. This ensures that any label inversion error is automatically corrected without manual intervention.


        \begin{equation}
         \text{acc} =  \frac{1}{n} \sum_{i=1}^n \mathbb{I}(Y_{\text{true}}^{(i)} = Y_{\text{pred}}^{(i)})
         \end{equation}

         \begin{equation}
         \text{optimal}\_{\text{acc}} =  \frac{1}{n} \sum_{i=1}^n \mathbb{I}(Y_{\text{true}}^{(i)} = 1-Y_{\text{pred}}^{(i)})
         \end{equation}

        \begin{equation}
            Y_{\text{final}} = 
            \begin{cases}
                1 - Y_{\text{pred}} & \text{if } \text{acc}>\text{optimal\_acc} \\
               Y_{\text{pred}} & otherwise.
            \end{cases}
            \label{eq:anomaly}
        \end{equation}

        

\begin{algorithm}                     
\SetKwInOut{Input}{Input}
\SetKwInOut{Output}{Output}

\Input{
  $y_{\text{true}}$: True labels (array of shape [n\_samples])\\
  $y_{\text{pred}}$: Predicted labels (array of shape [n\_samples])\\
  $n_{\text{classes}}$: Number of unique classes (integer)
}
\Output{
  $y_{\text{corrected}}$: Corrected labels after alignment\\
  $\text{final\_accuracy}$: Best achieved accuracy after correction
}
\BlankLine

  // Calculate initial accuracy \\
  $\text{accuracy} \leftarrow \text{accuracy\_score}(y_{\text{true}}, y_{\text{pred}})$ \\
  \BlankLine
  
  // Binary classification case (special handling) \\
  \If{$n_{\text{classes}} == 2$}{
    $y_{\text{inverted}} \leftarrow 1 - y_{\text{pred}}$ \\
    $\text{inverted\_accuracy} \leftarrow \text{accuracy\_score}(y_{\text{true}}, y_{\text{inverted}})$ \\
    \BlankLine
    
    \If{$\text{inverted\_accuracy} > \text{accuracy}$}{
      \Return{$y_{\text{inverted}}, \text{inverted\_accuracy}$}
    }
    \Else{
      \Return{$y_{\text{pred}}, \text{accuracy}$}
    }
  }
  \BlankLine
\caption{Label Alignment Algorithm for Anomaly Detection}
\label{alg:anomaly_detection}
\end{algorithm}

In Algorithm~\ref{alg:anomaly_detection}, we give the actual sublabels, predicted labels, and the number of classes as input to the algorithm, and it returns the optimized labels and corrected accuracy as output. At the beginning of the algorithm, the accuracy of the initial predicted state is calculated. The accuracy of the predicted inverse state for two classes (anomaly detection) is calculated, and if the improved accuracy is better, then the inverse of the labeled model is considered as the labeled model.

    \begin{figure}
        \centering
        \includegraphics[width=0.7\linewidth]{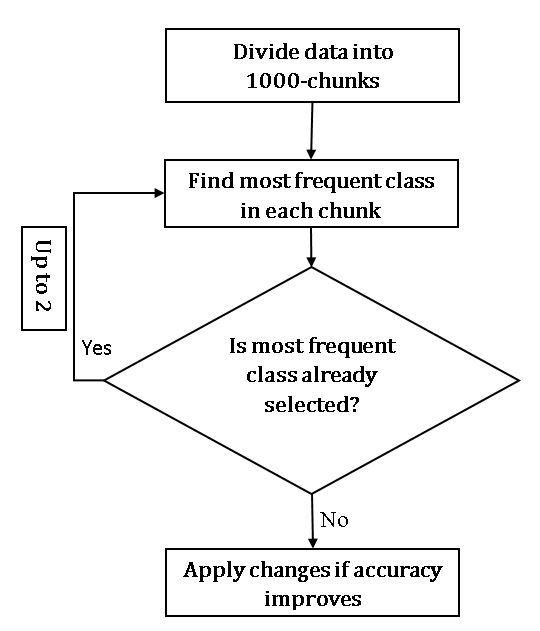}
        \caption{Label Assignment.}
        \label{fig:label_assignment}
    \end{figure}

\subsubsection{Device Identification (Multi-class Case)} 
In multi-class device identification with 11 classes, exhaustive label permutation is infeasible due to the large number of states. Instead, we apply a frequency-based label alignment strategy, with each step linked to its corresponding lines in Algorithm~\ref{alg:device_identification}:

\begin{itemize}
    \item Dominant class detection (Lines 4-9): Partition predictions into batches of 1,000 samples per true device class and select the most frequent predicted label as the dominant class.

    \item Conflict resolution (Lines 10-21): If a dominant class is already assigned, select the next most frequent unused class in a two-stage process until all classes are uniquely assigned.

    \item Label remapping and validation (Lines 22-29): Remap predicted labels to match the original class IDs and compute the corrected accuracy. If the corrected accuracy exceeds the initial value, we select the optimized classes as the predicted class.

\end{itemize}

Figure~\ref{fig:label_assignment} shows how the following algorithm works.



\begin{algorithm}
\caption{Label Alignment Algorithm for Device Identification}
\SetKwInOut{Input}{Input}
\SetKwInOut{Output}{Output}

\Input{
  $y_{\text{true}}$: True labels (array of shape [n\_samples])\\
  $y_{\text{pred}}$: Predicted labels (array of shape [n\_samples])\\
  $n_{\text{classes}}$: Number of unique classes (integer)
}
\Output{
  $y_{\text{corrected}}$: Corrected labels after alignment\\
  $\text{final\_accuracy}$: Best achieved accuracy after correction
}
\BlankLine

    // Step 1: Find dominant predicted class per true class segment \\
    $\text{dominant\_classes} \leftarrow \text{array of size } n_{\text{classes}} \text{ initialized with } -1$ \\
    $\text{class\_distributions} \leftarrow \text{empty list}$ \\
    \BlankLine
    
    \For{$i \leftarrow 0$ \KwTo $n_{\text{classes}} - 1$}{
      $\text{segment} \leftarrow y_{\text{pred}}[i \times 1000 : (i + 1) \times 1000]$ \\
      $\text{top\_class} \leftarrow \text{most\_common\_class}(\text{segment})$ \\
      $\text{\begin{small}class\_distributions.append\end{small}}(\text{\begin{small}frequency\_distribution\end{small}}(\text{\begin{small}segment\end{small}}))$ \\
      
      \If{$\text{top\_class} \notin \text{dominant\_classes}$}{
        $\text{dominant\_classes}[i] \leftarrow \text{top\_class}$ \\
      }
    }
    // Step 2: Handle unassigned classes \\
    \For{$i \leftarrow 0$ \KwTo $n_{\text{classes}} - 1$}{
      \If{$\text{dominant\_classes}[i] == -1$}{
        \For{$(\text{class, freq}) \in \text{class\_distributions}[i]$}{
          \If{$\text{class} \notin \text{dominant\_classes}$}{
            $\text{dominant\_classes}[i] \leftarrow \text{class}$ \\
            break
          }
        }
      }
    }
    // Step 3: Fallback assignment \\
    \For{$i \leftarrow 0$ \KwTo $n_{\text{classes}} - 1$}{
      \If{$\text{dominant\_classes}[i] == -1$}{
        \For{$\text{class} \leftarrow 0$ \KwTo $n_{\text{classes}} - 1$}{
          \If{$\text{class} \notin \text{dominant\_classes}$}{
            $\text{dominant\_classes}[i] \leftarrow \text{class}$ \\
            break
          }
        }
      }
    }
    %
    %
    // Step 4: Remap labels \\
    $\text{original\_classes} \leftarrow [0, 1, \dots, n_{\text{classes}} - 1]$ \\
    $y_{\text{remapped}} \leftarrow \text{copy}(y_{\text{pred}})$ \\
    
    \For{$i \leftarrow 0$ \KwTo $n_{\text{classes}} - 1$}{
      \If{$\text{dominant\_classes}[i] \neq \text{original\_classes}[i]$}{
        // Temporary offset to avoid conflicts \\
        $y_{\text{remapped}}[y_{\text{pred}} == \text{dominant\_classes}[i]] \leftarrow n_{\text{classes}} + i$ \\
        $y_{\text{remapped}}[y_{\text{pred}} == \text{original\_classes}[i]] \leftarrow \text{dominant\_classes}[i]$ \\
      }
    }
    
    // Calculate corrected accuracy \\
    $\text{corrected\_accuracy} \leftarrow \text{accuracy\_score}(y_{\text{true}}, y_{\text{remapped}})$ \\
    \BlankLine
    
    \If{$\text{corrected\_accuracy} > \text{accuracy}$}{
      \Return{$y_{\text{remapped}}, \text{corrected\_accuracy}$}
    }
    \Else{
      \Return{$y_{\text{pred}}, \text{accuracy}$}
    }

\label{alg:device_identification}
\end{algorithm}


\begin{figure*}[ht]
    \centering
    \begin{minipage}[b]{0.48\textwidth}
        \includegraphics[width=\textwidth]{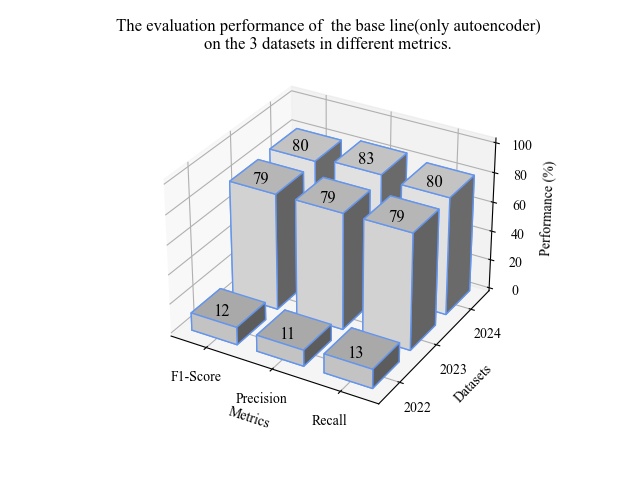}
       
        \begin{center}
             \textbf{(a) Baseline}
        \end{center}
        \label{fig:1a}
    \end{minipage}
    \hfill 
    \begin{minipage}[b]{0.48\textwidth}
        \includegraphics[width=\textwidth]{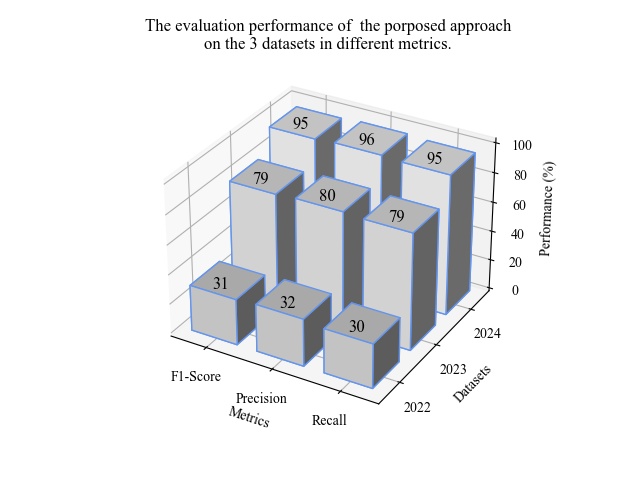}
        \begin{center}
            \textbf{(b) Proposed approach}
        \end{center}
        \label{fig:1b}
    \end{minipage}
    \caption{(a) Baseline autoencoder + K-means vs. (b) Proposed federated autoencoder + K-means; comparison of F1, Precision, Recall across CICIoT2022/2023/CICIoT-DIAD 2024}
    \label{fig:1}
\end{figure*}

The comparison between the two graphs in Figure~\ref{fig:1} highlights the performance improvements achieved by the proposed method over the baseline. Both graphs are three-dimensional, with the first axis representing the datasets (2022, 2023, and 2024), the second axis denoting the evaluation metrics (F1-score, Precision, and Recall), and the third axis showing the corresponding performance values. The left graph illustrates the baseline performance using only the autoencoder and K-means, while the right graph presents the final performance of the proposed federated autoencoder with K-means, reporting the best results across all training rounds. In the 2022 dataset, all evaluation metrics demonstrate clear improvements. For the 2023 dataset, the gains are more modest. However, in the 2024 dataset, performance improves significantly, with the F1-score increasing by approximately 15\%, underscoring the effectiveness of the proposed approach. These results suggest that the method is particularly effective on more recent and complex datasets, where feature heterogeneity poses greater challenges.


    
These findings clearly show that using heterogeneous networks and federated learning, rather than traditional methods, can lead to better results.
    \begin{figure}
        \centering
        \includegraphics[width=1\linewidth]{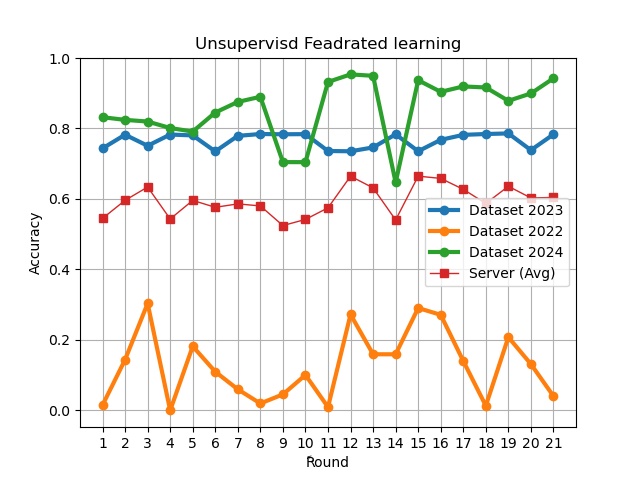}
        \caption{Comparison of Datasets Rounds}
        \label{fig:acc_rounds}
    \end{figure}

In Figure~\ref{fig:acc_rounds}, the accuracy is shown in 21 rounds for the client and the server. Each client is a dataset with a different number and label of features that is trained for two epochs. The average of the weights of the common layer is placed on the server. The clients are evaluated with it, and the average results based on the number of samples from the datasets are shown in this graph. Notably, the 2024 dataset, being the most recent and complex, shows the most substantial performance gains.


\subsection{Global Model Accuracy Aggregation}
Equation ~\ref{eq:update_accuracy} defines the accuracy of the global model as a weighted average of the local accuracies of all participants in the federated learning system. {$n$} is the total number of clients. {$test\_size\_{c}$} refers to the number of test samples evaluated by client c. {$accuracy\_{c}$} represents the local accuracy of the model trained by the client c.
\begin{equation}
    \text{server\_accuracy} =  \sum_{c=1}^{n}\frac{\text{test\_size}_c}{\sum_{i=1}^{n}\text{test\_size}_i}  \text{accuracy}_{c}
    \label{eq:update_accuracy}
\end{equation}

\subsection{\textcolor{black}{Evaluation Metrics}}


To assess model performance in device identification and anomaly detection tasks, we rely on the confusion matrix, which summarizes predictions against ground-truth labels and allows evaluation of different types of errors. From this, we derive four standard metrics: accuracy, precision, recall, and F1-score.





\begin{figure}
    \centering
    \includegraphics[width=1\linewidth]{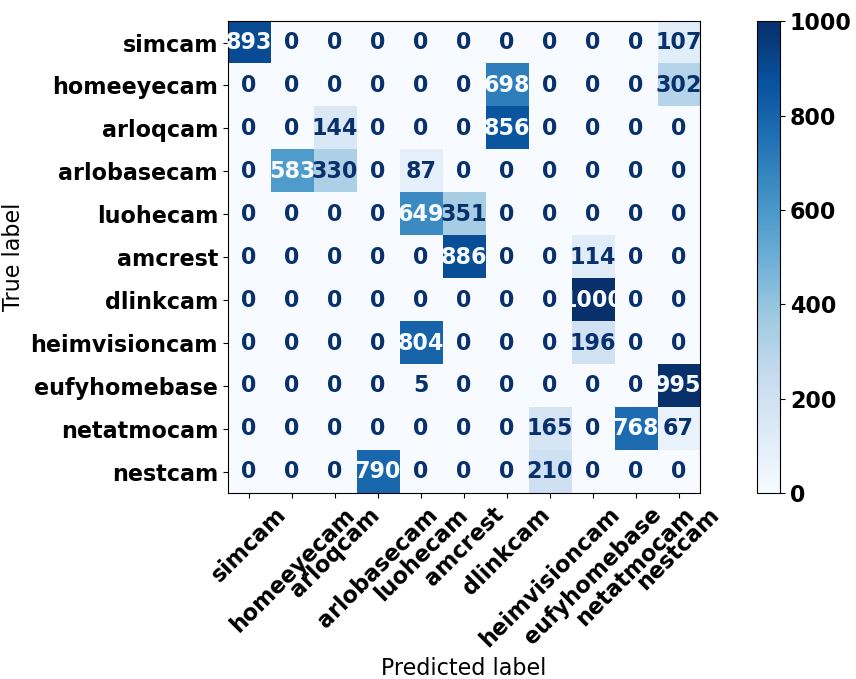}
    \caption{Confusion Matrix of dataset CICIoT2022(client 1) in round 3.}
    \label{fig:confusion2022}
\end{figure}
\begin{figure}
    \centering
    \includegraphics[width=0.8\linewidth]{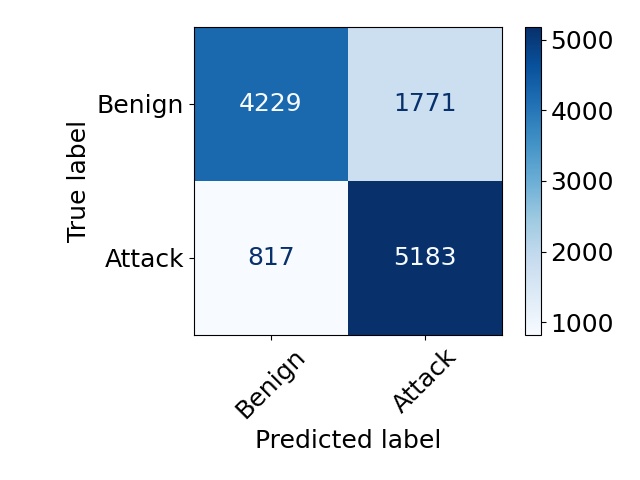}
    \caption{Confusion Matrix of dataset CICIoT2023(client 2) in round 8.}
    \label{fig:confusion2023}
\end{figure}
\begin{figure}
    \centering
    \includegraphics[width=0.8\linewidth]{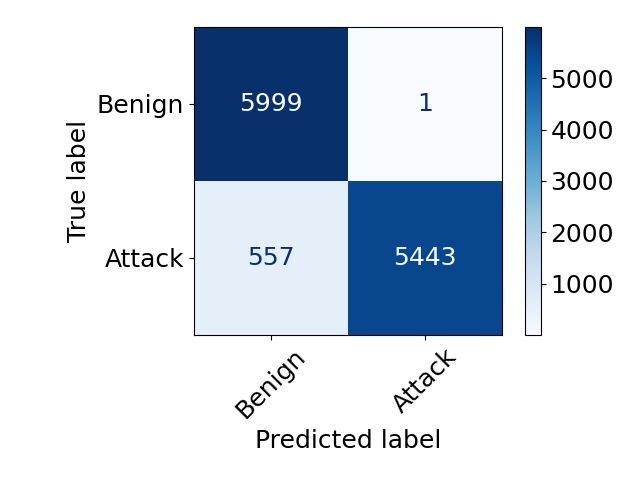}
    \caption{Confusion Matrix of dataset CICIoT2024(client 3) in round 12}
    \label{fig:confusion2024}
\end{figure}

During federated training, the best-performing models are saved, and their confusion matrices are reported. Figure ~\ref{fig:confusion2022} presents device identification results, while Figures~\ref{fig:confusion2023} and ~\ref{fig:confusion2024} illustrate the identification of anomalies.




\subsection{Evaluation Results}

    


Table~\ref{table:result} compares the performance of the proposed method with a baseline autoencoder model under identical conditions. The results clearly indicate that the proposed approach achieves notable improvements, particularly on the CICIoT-DIAD 2024 dataset, where an average gain of approximately 15\% is observed across the evaluation metrics. This improvement can be attributed to leveraging common features and shared structures within the extracted representations, enabling the model to generalize more effectively across heterogeneous data sources.

Based on the Tables~\ref{table:summary_atoencoder}, to construct the deep autoencoder model, we employ a combination of two sequential Artificial Neural Network (ANN) modules, comprising an encoder and a decoder. The encoder compresses the input into a low-dimensional latent representation, while the decoder reconstructs the original data from this embedding. Training is performed using Stochastic Gradient Descent (SGD), which updates parameters iteratively on mini-batches to accelerate convergence. To introduce non-linearity and enhance feature learning, the Rectified Linear Unit (ReLU) activation function is employed.

    \begin{table}
     \caption{Evaluation results on CICIoT Datasets.  }
        \label{table:result}
    \renewcommand{\arraystretch}{1.3}
        \centering
        \begin{tabular}{|p{1.4cm}|p{1cm}|p{1.2cm}|l|p{0.7cm}| p{0.7cm}| p{0.7cm} |}
            \hline
            \textbf{Method} & \textbf{Datasets} & \textbf{Accuracy} & \textbf{F1-score} & \textbf{Round} & \textbf{Epoch}\\
            \hline
            Autoencoder (baseline) & CICIoT 2024 & 0.8008 & 0.7952& - & 50\\ \hline
             Autoencoder (baseline) & CICIoT 2023 & \textbf{0.7918} & \textbf{0.7916} & - & 50\\ \hline
             Autoencoder (baseline) & CICIoT 2022 & 0.1295 & 0.1163 & - & 50\\
            \hline
              & CICIoT  2024 & \textbf{0.9535} & \textbf{0.9574}& 12 & 2\\
             Proposed Method & CICIoT  2023 & 0.7844 & 0.7857& 8 & 2\\
             & CICIoT  2022 & \textbf{0.30364} & \textbf{0.3064} & 3 & 2\\
            \hline
        \end{tabular}
    \end{table}


        
    

%% file: Explainability.tex
\label{sec:Explainability}


To provide interpretability, we employ SHAP (SHapley Additive exPlanations) to analyze the contribution of individual features to model predictions. Our primary objective was to minimize anomaly detection loss, which makes deriving meaningful explanations particularly challenging.

\begin{figure}
    \centering
    \includegraphics[width=1\linewidth]{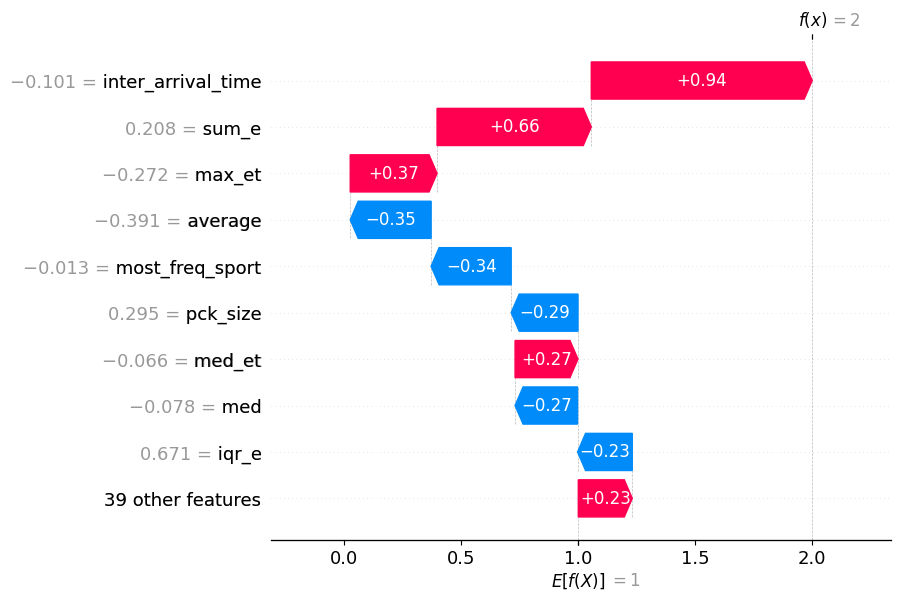}
    \caption{In the IoT 2022 dataset, showing the impact of different features on the target variable using SHAP}
    \label{fig:shape2022}
\end{figure}

\begin{figure}
    \centering
    \includegraphics[width=1\linewidth]{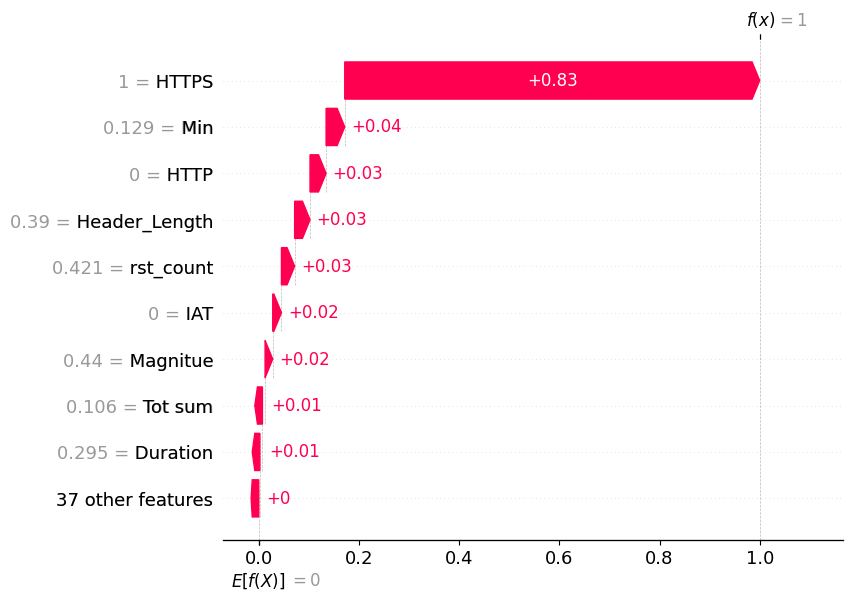}
    \caption{In the IoT 2023 dataset, analyzing the distribution of feature values using SHAP}
    \label{fig:shape2023}
\end{figure}

\begin{figure}
    \centering
    \includegraphics[width=1\linewidth]{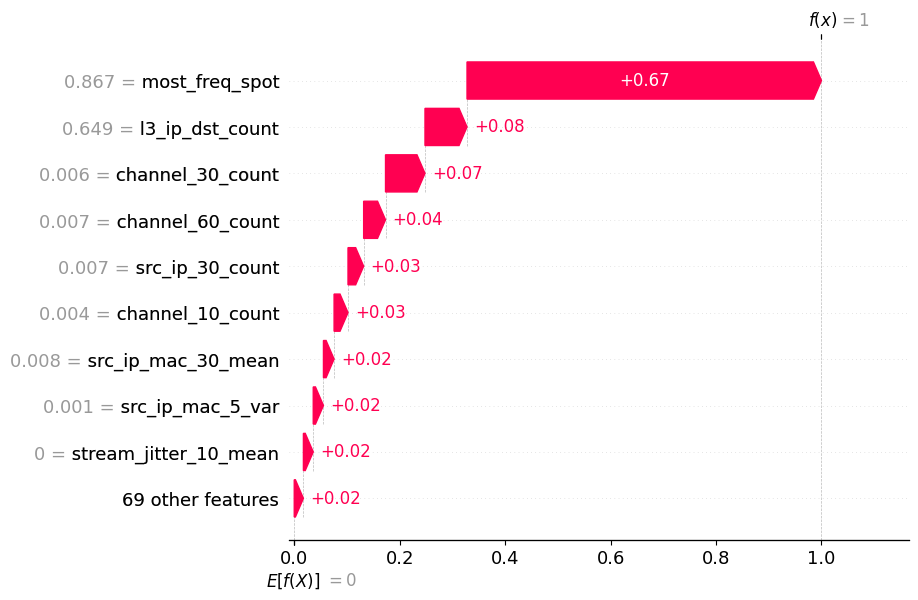}
    \caption{In the IoT-DIAD 2024 dataset, analyzing the distribution of feature values using SHAP}
    \label{fig:shape2024}
\end{figure}

Figures 13–15 illustrate SHAP values obtained from client models at Round 12. In Figure~\ref{fig:shape2022}, the change in the attribute when the target class changes from one device to another, and in other Figures~\ref{fig:shape2023} and ~\ref{fig:shape2024}, the amount of change in the attribute when the target class changes from a building to an attack, one classified as benign, and the other as an attack. In these plots, features are listed along the y-axis, with their contributions shown on the x-axis. Pink bars indicate a positive contribution toward predicting an attack, while blue bars indicate a negative contribution to predicting benign samples. The length of each bar reflects the relative importance of the feature. This figure demonstrates which feature values differed between the benign and attack samples, leading the model to predict an attack. 


%% file: conclusion.tex
\label{sec:summary}

Heterogeneity in IoT environments—stemming from diverse devices, feature spaces, and operating conditions—poses significant challenges for centralized anomaly detection and privacy-preserving learning. To address these challenges, this paper proposed an unsupervised federated learning (FL) framework specifically designed to integrate heterogeneous clients without sharing raw data. The framework achieves this by (i) aggregating only common-dimension layers across clients, (ii) dynamically aligning local and global model weights, and (iii) preserving dataset-specific features through local fine-tuning. Latent representations learned via deep autoencoders are clustered using K-means, supported by tailored label-alignment strategies for both binary anomaly detection and multi-class device identification. Furthermore, SHAP-based explainability is employed to interpret model decisions and validate the contribution of shared features.

Extensive experiments on CICIoT2022, CICIoT2023, and CICIoT-DIAD 2024 demonstrate the effectiveness of the proposed framework. By jointly leveraging shared and unique feature representations, the heterogeneity-aware FL approach consistently outperforms a centralized autoencoder baseline on CICIoT2022 and CICIoT-DIAD 2024, while maintaining comparable performance on CICIoT2023. Notably, on CICIoT-DIAD 2024, the proposed method achieves an improvement of approximately 15\% in F1-score, along with stable convergence over 21 federated rounds. The introduced label-alignment procedures enable fair evaluation of unsupervised clustering outcomes, and SHAP analyses confirm that shared features play a dominant role in driving the observed performance gains. 
To answer the Research Questions introduced in Section \ref{sec:introduction}, \textbf{RQ1: Federated learning under feature heterogeneity;} The experimental results show that heterogeneous and partially overlapping feature spaces can be effectively integrated within the proposed federated learning framework. By restricting aggregation to common layers and complementing it with local fine-tuning, clients with diverse feature sets can collaboratively construct a robust global model without performance degradation. Improvements on CICIoT2022 and CICIoT-DIAD 2024, together with stable convergence behavior, demonstrate that cross-dataset integration is not only feasible but beneficial when the shared feature subset is informative.

\textbf{RQ2: Detection effectiveness and explainability of federated models.}
The learned global federated model effectively detects device-level anomalies across heterogeneous datasets, with the strongest gains observed when meaningful feature overlap exists among clients. SHAP-based explainability provides consistent and interpretable insights into model behavior, revealing that the most influential predictors largely coincide with the shared feature subset. This alignment between explainability results and model design validates the framework’s ability to deliver both high detection performance and transparent decision-making.

\textbf{Limitations:} The proposed approach relies on a non-trivial overlap of shared features across clients; its performance may degrade as the extent of overlap decreases. The use of K-means clustering provides a simple decision rule but does not capture temporal dependencies in the data. Additionally, we did not quantify communication or energy costs, nor did we incorporate privacy-preserving mechanisms beyond maintaining data locality.

\textbf{Future Work:} Future research directions include:
(i) \textit{Learned representation alignment}, using contrastive learning, self-supervised objectives, or knowledge distillation to reduce reliance on manual or shared features;
(ii) \textit{Personalized or clustered federated learning} to better accommodate highly non-IID clients;
(iii) \textit{Temporal modeling}, e.g., sequence autoencoders or transformers, to capture flow- or session-level dynamics;
(iv) \textit{Robust and privacy-preserving aggregation}, including differential privacy or secure multi-party computation; and
(v) \textit{Systems-level evaluation} of latency, bandwidth, and energy consumption at larger scales.

Overall, the results demonstrate that leveraging shared features across complementary datasets, while preserving client-specific structure, enables unsupervised federated learning to achieve stronger anomaly detection performance in decentralized, heterogeneous IoT environments. This highlights the potential of the proposed approach to balance accuracy, interpretability, and privacy in real-world IoT applications.
